# Developing trustworthy AI applications with foundation models

**Authors**

Michael Mock [1]
Sebastian Schmidt [1]
Felix Müller [2, 1]
Rebekka Görge [1]
Anna Schmitz [1]
Elena Haedecke [2, 1]
Angelika Voss [1]
Dirk Hecker [1]
Maximilian Poretschkin [1, 2]

[1] Fraunhofer Institute for Intelligent Analysis and Information Systems IAIS Sankt Augustin, Germany

[2] University of Bonn, Bonn, Germany

April 2024 (translation of German version published in January 2024)

## Abstract

The trustworthiness of AI applications has been the subject of recent research and is also addressed in the EU's recently adopted AI Regulation. The currently emerging foundation models in the field of text, speech and image processing offer completely new possibilities for developing AI applications. This whitepaper shows how the trustworthiness of an AI application developed with foundation models can be evaluated and ensured. For this purpose, the application-specific, risk-based approach for testing and ensuring the trustworthiness of AI applications, as developed in the "AI Assessment Catalog – Guideline for Trustworthy Artificial Intelligence" by Fraunhofer IAIS, is transferred to the context of foundation models. Special consideration is given to the fact that specific risks of foundation models can have an impact on the AI application and must also be taken into account when checking trustworthiness.

Chapter 1 of the white paper explains the fundamental relationship between foundation models and AI applications based on them in terms of trustworthiness. Chapter 2 provides an introduction to the technical construction of foundation models and Chapter 3 shows how AI applications can be developed based on them. Chapter 4 provides an overview of the resulting risks regarding trustworthiness. Chapter 5 shows which requirements for AI applications and foundation models are to be expected according to the draft of the European Union's AI Regulation and Chapter 6 finally shows the system and procedure for meeting trustworthiness requirements.

# Contents

# List of figures

# Executive Summary

The trustworthiness of AI applications has been the subject of recent research and is also addressed in the EU's recently adopted AI Regulation. The currently emerging foundation models in the field of text, speech and image processing offer completely new possibilities for developing AI applications. This whitepaper shows how the trustworthiness of an AI application developed with foundation models can be evaluated and ensured. For this purpose, the application-specific, risk-based approach for testing and ensuring the trustworthiness of AI applications, as developed in the "AI Assessment Catalog – Guideline for Trustworthy Artificial Intelligence" by Fraunhofer IAIS, is transferred to the context of foundation models. Special consideration is given to the fact that specific risks of foundation models can have an impact on the AI application and must also be taken into account when checking trustworthiness.

Chapter 1 of the white paper explains the fundamental relationship between foundation models and AI applications based on them in terms of trustworthiness. Chapter 2 provides an introduction to the technical construction of foundation models and Chapter 3 shows how AI applications can be developed based on them. Chapter 4 provides an overview of the resulting risks regarding trustworthiness. Chapter 5 shows which requirements for AI applications and foundation models are to be expected according to the draft of the European Union's AI Regulation and Chapter 6 finally shows the system and procedure for meeting trustworthiness requirements.

# 1 Introduction

The emergence of foundation models has ushered in a new era in the development of artificial intelligence (AI). These models are trained on large multimodal data sets and can be adapted to a large number of tasks by fine-tuning, for example. They therefore differ fundamentally from "conventional" machine learning (ML) applications, which are usually trained for a specific task. The capabilities and functionalities of foundation models are therefore drastically changing the way AI applications are developed and used. In particular, they enable a variety of applications that previously seemed difficult to realize with conventional methods, such as realistic-looking conversations on complex topics with a chatbot over a longer period of time - i.e. passing the Turing test - or a detailed and even interpretive description of sophisticated image content. Another special feature of the foundation models is that they can be adapted to a wide variety of tasks without much effort or can even be used directly without adaptation. For example, anyone without prior knowledge of AI can use ChatGPT's prompting function to create templates for speech manuscripts, emails or project reports. While until the appearance of foundation models, it was mainly repetitive tasks, for example in the context of process automation, that were influenced and even completely taken over by AI applications, this now also applies to creative professions, for example in media design. The resulting economic potential is enormous and is estimated at between 2.4 and 4.4 trillion US dollars [CH23a]. As a result, AI supply and value chains are subject to enormous changes. At the same time, regulatory efforts for artificial intelligence are being made in all major economic areas worldwide, above all in the European Union.[1] A key component of the European AI Regulation (EU AI Act) [EU23] is a conformity assessment of high-risk systems before they are placed on the market. AI-specific test procedures are a necessary prerequisite for carrying out such AI conformity assessments. One challenge here is that AI applications that are implemented using machine learning are not explicitly programmed, but the functionality is learned from large data sets. As a result, conventional procedures for quality assurance and quality verification of software fall short. An important approach at this point is to extend the concept of an operational design domain known from the automotive industry to other contexts and domains in such a way that the input space is provided with a semantic structure that allows a systematic search for weak points in the AI application and can serve as the basis for structured validation arguments and safety cases. Another challenge is that AI applications are generally subject to complex value chains. Important participants in the value chain include hardware providers, data providers, framework providers and providers of basic AI services (such as OCRs) and toolboxes. The question of the transfer points of responsibility for individual components of an AI application along the value chain - for example, how a non-accessible OCR is to be assessed as part of an AI application to be tested - is still often a challenge in practice. At the same time, a number of promising approaches for assessing AI applications have been published in the recent past [PO21, LE21, NI23, AI23, MO23], which are already the subject of various standardization activities. In addition, the formation of an AI testing industry can currently be observed.

In the case of assessing AI applications that are based on foundation models or include them as a component, the challenges just described are exacerbated once again: On the one hand, this is due to the fact that it is practically impossible to construct an analog to the concept of the operational design domain known from the automotive domain, i.e. a systematic description of the possible input space for foundation models. Instead of targeted tests, benchmarks are therefore used to systematically compare the performance of different foundation models for specific task types. Furthermore, conventional test methods include the context in which the AI application is used, for example for a risk analysis of possible damage that could potentially be caused by the system. In view of the wide range of possible uses of foundation models, however, such a context of use is also difficult to define, which led to the discussion during the finalization of the draft AI Regulation as to whether foundation models should therefore be classified as high-risk systems. Finally, foundation models further exacerbate the value chain problem, as the inclusion of such a foundation model entails a completely different level of complexity than the use of an OCR as a third-party component. At the same time, a particularly large number of AI services can be based on foundation models. A review of AI applications based on foundation models must therefore take two perspectives into account: Firstly, the question of the correctness and trustworthiness of the foundation model itself, and secondly, the question of whether an AI service or system based on it correctly implements the required downstream task. In view of the market power of the providers of such foundation models, an answer to the first question presupposes corresponding pressure from regulatory or market requirements. The second question is particularly pressing, namely the correctness of services or systems that are based on foundation models. This is because such services and systems are so quick and easy to implement that an accompanying quality assurance process is not always to be expected in practice. This in turn increases the probability of errors occurring.

---

[1]   For example, a "Code of Conduct" for artificial intelligence was published by the G7 in October 2023 [BMDV23].

This white paper starts here and shows a systematic approach to assessing AI applications that are implemented with foundation models. It presents a risk-based system that can be used to assess and ensure the trustworthiness of such AI applications. In addition, it provides an overview of the risks arising from the use of foundation models, but also highlights new possibilities (for example in the development of tests) for the development of trustworthy AI applications.

In order to motivate the connection between AI applications, trustworthiness and foundation models, Figure 1 first establishes a link to classic software development and considers the special case in which a foundation model generates software code. This functionality is currently already offered as an assistance function, for example by the Github Co-Pilot [NG22], and could be further automated in the future.

Test-based development is the state of the art in software development. Software tests use a number of different test cases to check whether the application works according to its specification. As part of so-called agile software development, these tests are even carried out continuously during software development and help to quickly create high-quality software together, even in large development teams.

If such a team no longer wanted to have every single line of code written by a developer, but also wanted to use code[2] generated by a foundation model, there would be no reason to dispense with the tests. On the contrary, the presence of tests would show that it is also possible to use code generated by a foundation model. As in the case of classic software development, the tests would thus justify confidence in the functionality of the application.

This situation can also be transferred to the case of AI application development, as outlined in Figure 2. AI applications

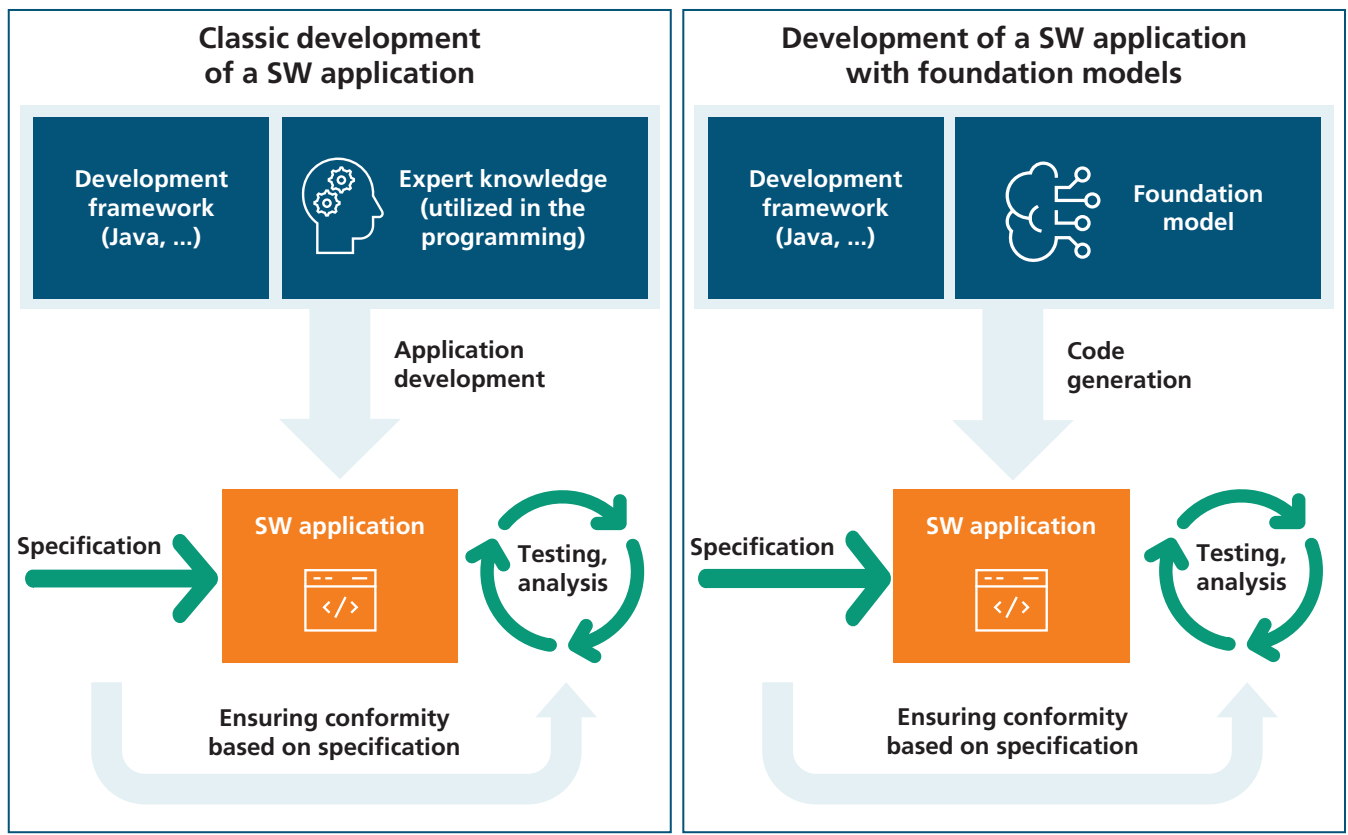

*Figure 1: Specification and testing in the development of a software application (SW application) always take place in relation to the specific use case. Confidence in the conformity of the application is achieved through the classic software test procedures, which are carried out based on the specification. On the left, the process for the classic development of a software application is shown, while on the right, a foundation model is used to generate the code for the software application.*

---

**2**   One example of this is the AI-based application "GitHub-Copilot", which supports programmers by generating and auto-completing code [GHC23].

are known to be subject to certain AI-specific risks, which can affect different goals, stakeholders and properties of the application and are sometimes referred to as dimensions of trustworthiness.[3] The trustworthiness of an AI application is particularly associated with the control of these risks and requirements for trustworthiness extend across the various dimensions, for example requirements for sufficient reliability or fairness. Since the application no longer works according to an algorithm specified by humans, but with a "trained model", the question arises in particular as to whether it also generalizes to the real application domain, i.e. whether it actually works on new input data that was not present in the training data.

How high these requirements are in the various dimensions depends on the use case. For example, a safety-critical application, such as person recognition in automated driving, will have to meet higher reliability requirements than AI-based software that makes purchase recommendations in a web store. Some AI applications are subject to high fairness requirements, for example in the automated granting of loans, while other applications typically have lower fairness requirements, such as quality assurance in automated manufacturing processes. To make matters worse, there are trade-offs between the various dimensions of trustworthiness. The weighing up of different requirements is based on the purpose of the AI application and does not result from the technology used to develop the application. The current draft of the EU Regulation on AI [EU23] also classifies AI applications into risk levels depending on the purpose of the application, and not according to technology levels.

As in traditional software development, tests are used in AI applications to verify the application-specific requirements. Depending on the requirements profile, these tests not only measure reliability, but also other properties of the application. The test data plays a central role here. It should cover the application domain as well as possible. Correct functioning of the AI application on the test data creates confidence in the functionality of the real AI application.

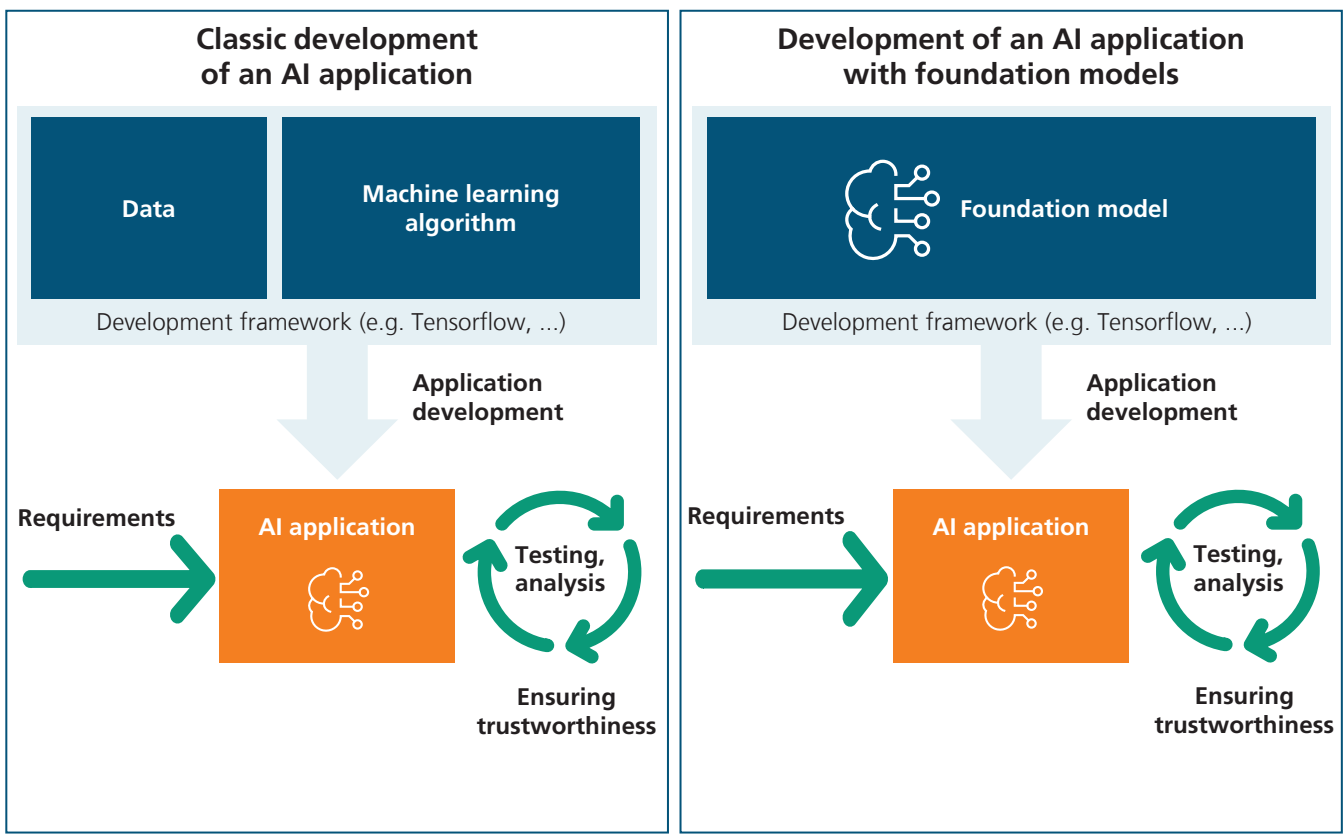

*Figure 2: The trustworthiness requirements and their assessment take place at the application level during the development of AI applications. The left part of the figure shows the classic development process compared to the right part, where foundation models are used for development.*

---

**3**  The "AI Assessment Catalog" from Fraunhofer IAIS distinguishes between six risk dimensions of trustworthiness [PO21].

Foundation models now offer a completely new basis for developing AI applications. Foundation models can, for example, be retrained for specialized applications with significantly less data (so-called fine-tuning) than would be necessary for a new development of the application. There are also new methods for developing AI applications (see Chapter 3) that use and call up foundation models as a kind of "library". However, as in traditional software development, both the requirements for the trustworthiness of the AI application and its verification are application-specific and require application-specific tests (and test data) in particular. In addition to the new possibilities for developing AI applications, foundation models also open new perspectives for testing AI applications by generating test data with foundation models.

This white paper provides an insight into the context and shows a practicable way to develop trustworthy AI applications with foundation models.

# 2 The basics of foundation models

What makes foundation models such a powerful building block from a technical perspective and how do they differ from previous AI models?

Basically, all AI models are trained on data in order to make the inherent knowledge in the data algorithmically usable. A classic distinction is made between "supervised learning" and "unsupervised learning", whereby "learning" in both cases refers to an optimization process based on stochastic laws (so-called machine learning). The two variants are shown schematically in Figure 3.

In the case of supervised learning shown in Figure 3, a model $\hat{f}$ tis trained to approximate a given target function $f$ as well as possible. The operating principle of the target function $f$ is unknown, otherwise it could have been programmed as a classic algorithm. Instead, the target function $f$ is only specified using examples (training data). A model for speech recognition, for example, would be trained with audio recordings of spoken example sentences, for which the associated texts are also available as so-called "labels" and together form the so-called training data. The learning algorithm iteratively adapts the model $\hat{f}$ to the training data so that the total error $|f\text{-}\hat{f}|$ on the training data is as small as possible. In abstract terms, this automatically extracts the previously only intrinsic knowledge about the causal relationship between the inputs and the output. The great advantage of this method is that the learned model $\hat{f}$ can also be applied directly to new data, for example to previously unknown (to the model) sound recordings. The disadvantage, however, is that the manual genera­tion of labeled training data sets often involves a great deal of effort that cannot be automated. This naturally limits the amount of usable training data.

The disadvantage of the need for labels for training is eliminat­ed in unsupervised learning from Figure 3, as there is no target function $f$ here. Nevertheless, a training process for the model $\hat{f}$ is also carried out here, which aims to describe the internal structures and any correlations in the data as well as possible.

Examples would be a shopping basket analysis (itemset mining), i.e. which objects are often bought together, group formation (clustering) in the data or graph mining algorithms. Also in unsupervised learning, $\hat{f}$ is usually determined itera­tively and the quality of the approximation to the structures actually presented in the data is measured using an error term. As a result, the $\hat{f}$ model conveys explicit knowledge that was previously only immanently present in the data. However, a model trained using unsupervised learning is usually descriptive in nature and therefore has the disadvantage that it cannot be applied to new data.

Foundation models are based on a newer methodology known as "self-supervised learning". As shown in Figure 4, a founda­tion model $\hat{f}$ is trained on (unlabeled) input data and can still be applied to new data.

The methodology of self-supervised learning is named after the fact that the target function $f$ generates the labels from the data itself, so that a model $\hat{f}$ that approximates the target function $f$ as well as possible can then be trained using the known supervised learning technique from Figure 3. The simplest example of such an objective function would be the "identity", which states that the input should be reproduced as correctly as possible. Other examples of simple objective functions are omitted words in a text, blackened sections in an image or predicting the next frame in a video sequence. Further transformations of the input are also possible, which must then be reproduced. Other complex target functions $f$ explicitly model relationships and correlations found in the data, similar to unsupervised learning. One example is the CLIP model [RA21], which learns the objective function whether an image can be found on a website together with a certain caption or not.

An important part of the learning process for $\hat{f}$ is that a mapping $\hat{f}_{emb}$ is created that maps the input data to so-called "embeddings", which are simply real-valued number vectors (typically at least of length 512). However, a large number of

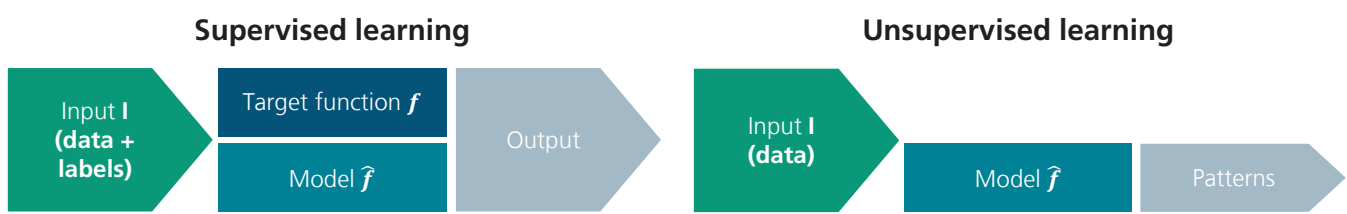

*Figure 3: Distinction between supervised and unsupervised learning.*

**Self-supervised learning**

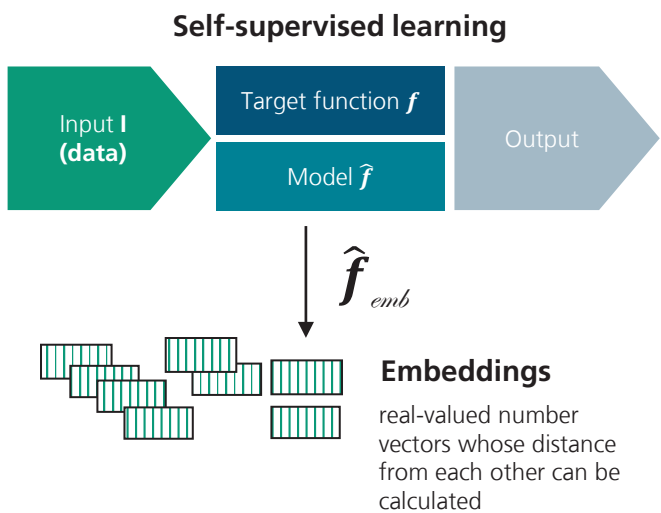

*Figure 4: Self-supervised learning.*

studies and experiments have shown that the values and distances of the embeddings have semantic meaning. A famous early example was able to that show for word embeddings the equation $\hat{f}_{emb}$("King") - $\hat{f}_{emb}$("Man") = $\hat{f}_{emb}$("Queen") holds [MSC13]. To a certain extent, the foundation model $\hat{f}$ with its associated embedding function $\hat{f}_{emb}$ can semantically correctly locate almost any input data in the space of embeddings so that other applications can immediately reuse it.

An example of such an application is one of the first generative image models DALL-E 2 [OA22], which can generate images matching a textual description of an image (e.g. "astronaut on a horse"). The application is based on the CLIP foundation model mentioned above. With its embedding function, CLIP can display both images and texts in the same embedding space. The challenging task for DALL-E 2 is now to generate an image B similar to a text T in this embedding space. DALL-E uses a two-stage process for this. First, based on the text embedding $\hat{f}_{emb}$(T), a plausible image embedding $\hat{f}_{emb}$(B) is generated in the vicinity of $\hat{f}_{emb}$(T) (DALL-E Prior model). In a second step, DALL-E generates an image in an iterative "diffusion process" that is as similar as possible to the proposed image embedding and thus also to the input text. It would also be possible to generate images directly based on $\hat{f}_{emb}$(T), but using the prior model improves the quality and variability of the generated images.

Generative language models go one step further: the embedding function is also used to learn a decoding function that can create suitable texts for given embeddings. This enables the language model to find the "best fitting" continuation for a given input from the training data, whereby the "best fitting" is, on the one hand, the continuation with the highest probability among all possible continuations. On the other hand, due to the rich semantic meaning of the embeddings, this continuation refers more to the meaning than to the pure word sequence of the input. It has been shown empirically [RA19] that, if the language model is sufficiently large, input texts such as "Translate from German to English" are also mapped semantically correct as the formulation of a task and thus the most likely continuation in terms of meaning is the translation of the subsequent input text into English. This means that the input text contains both the task to be performed and the input for this task, without having to be explicitly separated from each other in the word sequence. Implicitly, the foundation model has learned many different such tasks in the sense of multi-task learning through the self-supervised training process, the carefully selected target function and the semantically meaningful embedding space filled as a result. In supervised learning, each of these tasks would have had to be defined as part of a target function via labels. As in the CLIP model discussed above, the basic ability to semantically locate textual task descriptions correctly can also be combined with self-supervised learning methods in the field of image processing. One example of automatically learned "downstream" tasks is the generation of so-called synthetic training data by a foundation model. For example, the foundation model described in [ME23] marks the pedestrians appearing in an image based on text input such as "people, pedestrians"; in the case of text input such as "cars, wheels", it marks the wheels of the cars appearing in the image.

In summary, foundation models are characterized by the following properties:

- Foundation models are based on self-supervised learning and do not require manually generated labels.
- Foundation models can be trained on the largest possible amount of data because no labels are required.
- Foundation models have a very large number of internal parameters [AN23] - between 100 billion and more than 500 billion parameters, depending on the model. This allows them to generate (embedding) functions that map complex semantic properties of the real data in easily manageable mathematical values.
- Foundation models can process tasks formulated in inputs and predict the results for this task for an "actual input" given in the input. No explicit distinction is made in the input between the task and the actual input. They have learned this ability implicitly through self-monitored training [WE22a].

Ultimately, foundation models extract the knowledge inherent in the data and create functions that can be used to work with the data directly at the semantic level. This allows AI applications to be developed in new ways, for example without or with significantly less labeled data. The rapidly growing number of offerings and assistance systems that have already been brought to the market in a short space of time with the help of foundation models illustrates the enormous economic potential of this technology. In the first 6 months alone after the introduction of the plugin technology for ChatGPT, over 900 plugins have been developed and offered that the foundation model can call on demand, and there are "prompt markets" in which over 100,000 prompts are offered [WH23, ST23].

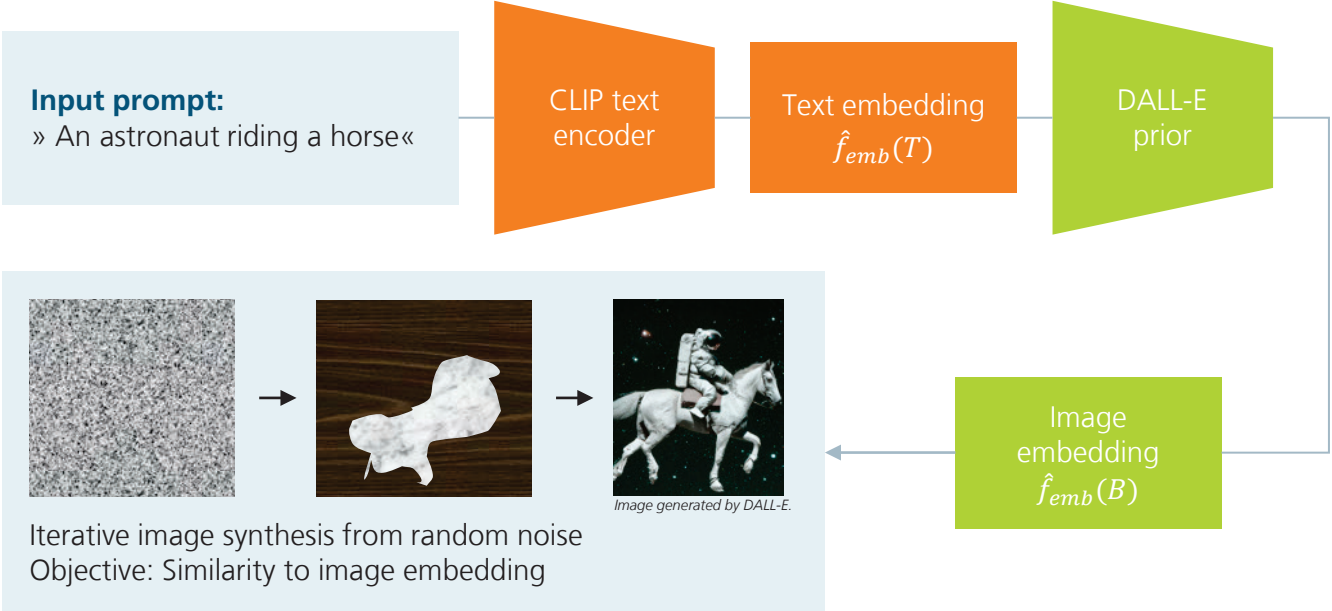

*Figure 5: Generation of semantically matching images for a prompt with DALL-E 2.*

# 3 From the foundation model to an AI application

By training on large amounts of data, foundation models learn fundamental knowledge that is useful for many domains and tasks. The specialization of a foundation model to the specific application domain and the actual task is an important option for using a foundation model for a specific AI application. Application-specific data and domain knowledge are needed to make the knowledge about semantic relationships implicit in the foundation model usable for specific AI applications. Due to the knowledge already available in the foundation model, this step is significantly simplified compared to developing an AI application without foundation models. For example, significantly less or no additional data is required. The following figure provides an overview of currently used options for application development with foundation models.

The various options are explained below, with the order roughly based on the amount of additional data required. It should be noted that the application development options presented are not just alternatives, but can also be combined.

## Programmed prompt engineering

**Zero-shot methods** are approaches that do not require any further data to create the application. For example, CLIP can be used to categorize images into predefined classes using the knowledge learned about images and captions without further training data. Each class is described via an input text (a prompt), for example "A picture of a dog", "A picture of a cat", "A picture of a house". The embedding function is then used to determine for which of these "prompts" the distance $f_{emb}$(prompt) to $f_{emb}$(picture) is minimal. The corresponding class is then output as the result of the application. Zero-shot methods can also be used in pure language models. Here, the relationship explained in Chapter 2 is exploited, in that both the task contained in an input and the actual input for the task are represented semantically in the embedding space. For example, the input "How old was Albert Einstein?" contains the task of determining the age of a person as well as the information to determine the person. Accordingly, the model can predict the answer as the "most probable continuation". In zero-shot methods, the actual task (or class sought in the example of CLIP) is formulated textually via a prompt, which is then mapped into the embedding space. It has been shown that the current foundation models react very sensitively to the concrete formulation of this prompt. As things stand today, it takes a lot of trial and error and experience with a foundation model to develop well-functioning prompts for a given application that define the task for the foundation model precisely enough. This process is often referred to as **prompt engineering**, as it is not classic development work, but rather an activity on the part of the user. In order to check which prompts are "good" or "good enough", the actual AI application is tested on test data sets in the zero-shot process. This means that additional, usually labeled data sets are also required for zero-shot procedures - even if only to test the application. In addition to prompt engineering for the development of AI applications, this method is of course used in the interactive use of generative AI. In this case, however, the question of the general validity of the prompts found for subsequent tasks is less important.

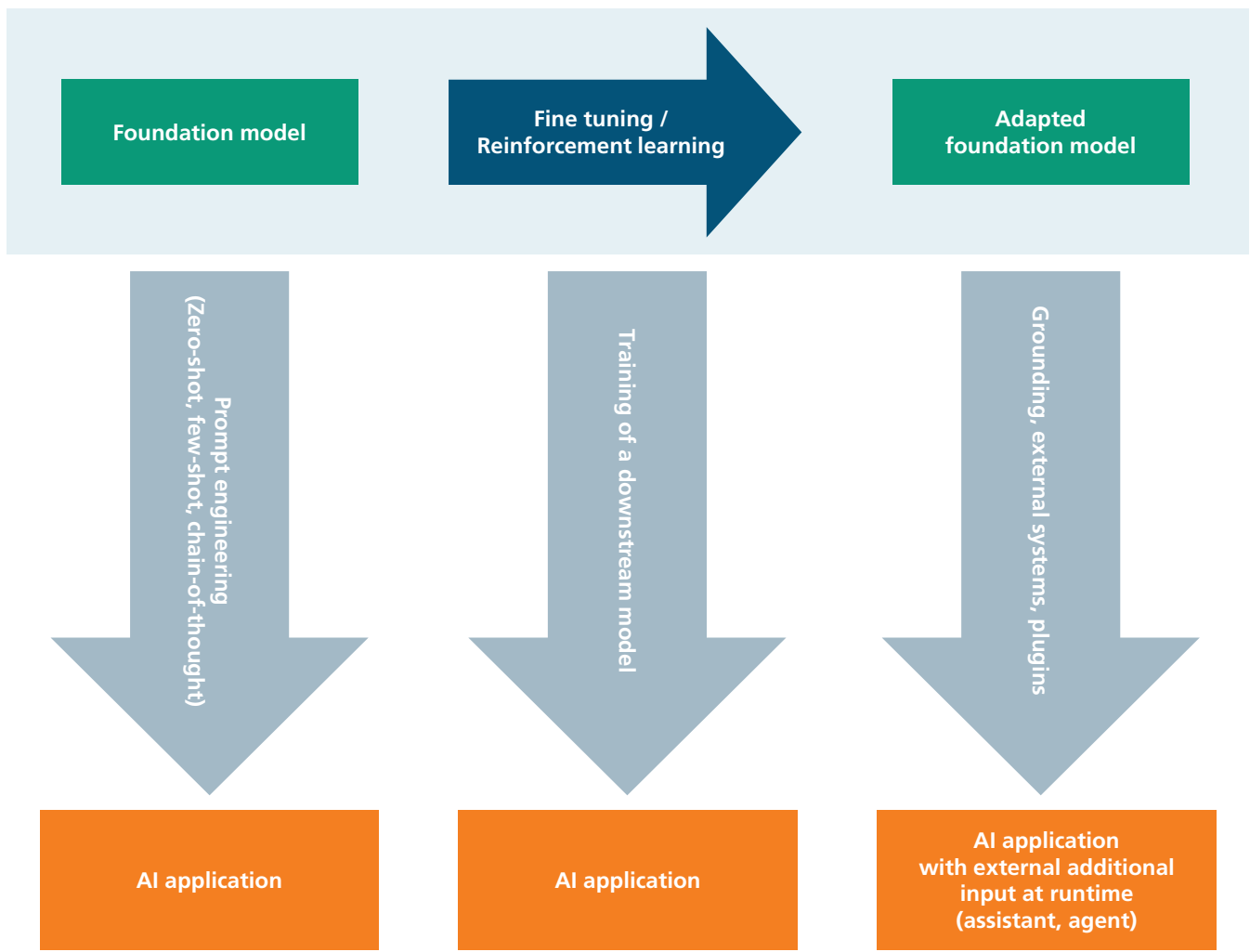

*Figure 6: Overview of methods for using foundation models to develop AI applications.*

A further development of the zero-shot method is known as **few-shot learning**. In the sense of prompt engineering, the task to be solved is specified - in addition to the task description and the actual input - by specifying additional, labeled data sets. This means that a fixed, small amount (e.g. 10 correct question-answer pairs) of labeled data from the domain of the application to be developed is given as context in the input. It has been shown empirically on large test data sets that this can significantly increase the accuracy of the output of foundation models. This method is particularly efficient when the inputs and possible outputs are highly standardized, for example when answering multiple-choice questions.

The term "learning" in "few-shot learning" is actually misleading, as the existing foundation model is not modified, i.e. no further training process takes place. Only the task to be fulfilled by the foundation model (see Chapter 2) is described in more detail by giving examples in the input. An example of "one-shot learning" would be the input

*"Translate from English to French: dog =>chien, cheese=>"*

in which a labeled example of the task to be performed is given. The expected output is therefore the translation of "cheese" into French, i.e. "fromage". Overall, the few-shot method is an important building block for a series of techniques that can be used to enrich the input of the foundation model with additional contextual information using a small amount of data.

The so-called **chain-of-thought** method [WE22] uses the technique of the few-shot method to not only describe the task to be solved by specifying correct question-answer pairs, but also to explicitly state the logical steps and conclusions for generating the correct answer as part of the input. The idea of the chain-of-thought method is thus to guide the model by providing examples in the input to explicitly state the underlying conclusions in addition to the answer. It is also possible to ask the model to specify the thought steps in the answer (e.g. by using the phrase "think step by step") without having to formulate the sub-steps as input. It has been shown that this can significantly increase the accuracy of the output of foundation models, for example when solving arithmetic text tasks on corresponding test data sets [SU22]. In principle, this procedure also opens the possibility of incorporating knowledge about the methodology of problem solving in an application domain into application development. However, there is still limited experience of the generalization capability in real application domains.

## Training of a downstream model

The application development methods described so far use no or only little labeled data and do not train new models or adapted foundation models. If more labeled data is available, further possibilities arise. A simple procedure is the **training of a downstream model**. The basic principle can be easily explained using the example of developing an application for image classification using CLIP. As described in the above section on zero-shot methods, the aim of the application is to assign a given image to one of several given classes (e.g. "dog", "cat", "house"). Labeled training data is available for training a supervised model. Instead of training this model $\hat{f}$ directly on the image data, the embedding function of CLIP is first applied to the images, i.e. the pairs ($f_{emb}$(image), label) are used as training data. Accordingly, the prediction for a new image is now obtained as $\hat{f}(f_{emb}$(image)). It could be shown on large test data sets that this method reduces both the required complexity of $\hat{f}$ and the amount of data required for training by orders of magnitude compared to existing deep neural networks for image classification that have been trained in a supervised manner from scratch, without having to accept any loss in the accuracy of the predictions.

## Fine-tuning and reinforcement learning through feedback

The procedure described for training a downstream model does not require any changes to the foundation model used and is therefore relatively easy to carry out. Fine-tuning is generally more complex in terms of the infrastructure used. The basic technique of **fine-tuning** has long been established, for example in the field of image processing, where pre-trained deep networks are used as a backbone, which have been trained on general image data for object recognition, for example. In fine-tuning, they are then trained further on special (labeled) training data of the application domain. In the same way, a foundation model can be further trained based on application-specific data and the internal parameters of the model can be optimized accordingly. This technique is also used by the manufacturers of foundation models in addition to self-supervised learning in order to train a language model to better understand and follow instructions in the prompt. Furthermore, the output for subsequent queries is also influenced and controlled by prompts executed in advance (so-called instruction fine-tuning). "Reinforcement learning by human feedback" is another method by which manufacturers try to train their model to produce responses preferred by humans, such as being polite and helpful and not giving out unauthorized information [GL22]. The term "alignment" is used in this context. In this method, a separate evaluation model is trained from human feedback and the foundation model is occasionally trained in such a way that it produces better-rated answers.

## Grounding and plugins

The application development techniques described so far are used in the design and development phase of the actual AI application. Accordingly, this results in an AI application specialized for one purpose and one domain of application, which should in principle be tested for its properties, especially with regard to the dimensions of trustworthiness, before it is put into operation. In addition, however, applications can also be developed that use the prompt engineering technique to obtain additional, application- and situation-dependent information from external sources during the runtime of the AI application, which ultimately makes the task to be solved by the foundation model as precise as possible in the specific case of the current query. In so-called **grounding** [MC23, NA21], the AI application is programmed in such a way that it performs queries to external systems at runtime, the results of which are then entered into the prompt according to an application-specific algorithm before the foundation model is called. In principle, there are no limits as to which external systems can be used. This can range from a simple search query in an Internet search engine to application-dependent, high-quality database systems and specialized document and knowledge databases. It is even possible to call up other foundation models.

In addition to using external systems, applications can also obtain information from long-term interaction with users and use this to enrich the input to the foundation model.

An even greater dynamic at runtime can be achieved using so-called **plugins**. As in grounding, a plugin can retrieve information via calling external systems. The foundation model itself generates the plugin call, which the surrounding software recognizes, executes and replaces with the result in the input text. In addition to the pure retrieval of information, actions (such as sending an email) can also be called via plugins. In this way, applications often referred to as "**autonomous agents**" can be created. These autonomous agents design the interaction with the environment dynamically from the runtime output of the underlying foundation model.

# 4 Risks of AI applications and foundation models

Various approaches and frameworks pursue the goal of making the trustworthiness of AI applications assessable, also with regard to the European AI Regulation, and providing guidelines that must be observed in the development and operation of trustworthy AI applications [PO21, LE21, VD22, OE22]. Typically, immanent AI risks, for example regarding the reliability or security of the application or with regard to possible discriminatory effects, are systematically examined in various dimensions of trustworthiness.

Even if the manufacturer of a foundation model has already taken extensive safeguarding measures through fine-tuning (see explanation in Chapter 3), the use of foundation models in AI applications does not, of course, render a systematic investigation of application-specific AI risks obsolete. However, further challenges may arise [BS23, LE23, WA23], which in turn have an impact on the AI application. Figure 7 summarizes specific risks in the use of foundation models along the dimensions of trustworthiness described in [CR19, PO20, PO21] and maps them to their significance in an AI application. This makes it clear that it is impossible to safeguard a foundation model for all conceivable application contexts, but that risks must always be considered in the context of the application. The following presentation of the dimensions of trustworthiness is based on the "AI Assessment Catalog - Guideline for Trustworthy Artificial Intelligence" [PO21] developed by Fraunhofer IAIS and describes the implications of using foundation models in AI applications for the individual risk dimensions.

## Fairness

The fairness dimension is intended to ensure that the AI application does not lead to unjustified discrimination and stereotyping. Typical causes of this are unbalanced (biased) training data or the statistical underrepresentation of groups of people, which can lead to a reduced quality of the AI application in relation to these groups. Since the foundation models are not trained on fully curated data, but on as much (all) data as possible, unequal distributions in the existing data are initially partially adopted. It is known that problems relating to the preference/disadvantage of gender-specific or ethnic groups in particular play a role here and can lead to **discrimination and stereotyping**. Examples include anti-Muslim completions in GPT-3 [AB21] or the image generation of DALL-E, which, before countermeasures were taken, produced predominantly male outputs under gender-neutral entries such as "heroic firefighter" or "CEO" [OB22]. Even if the manufacturers of foundation models explicitly disclose such fairness deficits on the one hand and counteract them with filtering techniques on the other, it should be noted that the use of a foundation model can exacerbate rather than mitigate the fairness problem. Tests in this regard at application level remain indispensable.

| Dimension of trustworthiness | Significance in the AI application | Specific risks that arise when using foundation models |
|---|---|---|
| Fairness | Does AI treat all those affected fairly? | - Stereotyping, discrimination |
| Autonomy & control | Is a self-determined, effective use of AI possible? | - Self-assured expression leads to blind trust<br>- Atrophy of rarely used skills (enfeeblement)<br>- Autonomous agents that develop unexpectedly<br>- Emotional dependencies |
| Transparency | Are the functioning and decisions of the AI comprehensible? | - Artificially generated content is difficult to recognize<br>- Invented justifications |
| Reliability | Does the AI work reliably? | - Lack of topicality<br>- Toxic and other prohibited content<br>- Hallucinations, misinformation, generation of code with errors<br>- Access to external systems (plugins) |
| Safety&Security | Is the AI safe and secure against attacks, accidents and errors? | - Disinformation, deepfakes, generation of malicious code<br>- Personalized fraud<br>- Attacks through unexpected or manipulative inputs |
| Data protection | Does the AI application protect sensitive information? | - Input or output of sensitive content<br>- Use of legally protected content |

*Figure 7: Specific Risks of foundation models in the dimensions of trustworthiness.*

## Autonomy and control

The fact that an AI application should not impair the autonomy of humans is already anchored in the basic ethical principles of the HLEG Commission [EU19], to which the planned AI Regulation is also oriented. The "AI Assessment Catalog" from Fraunhofer IAIS operationalizes this fundamental requirement at the level of the design and purpose of the AI application. The first step is to assess what degree of autonomy is appropriate for the purpose of the AI application. Then it is examined whether the human is adequately supported by the AI application and is given sufficient freedom of action in the interaction with the AI application. With regard to foundation models, this evaluation is of course only possible in the context of a specific derived AI application, as the purpose and the possibilities of interaction with humans are only defined here. In this sense, this dimension remains relevant to the assessment of trustworthiness even when using foundation models. It should also be noted that the known risks of "blind trust" and "emotional dependency" can increase, especially when language models are expressed in a human-like, often self-assured manner. In addition, the use of more powerful foundation models increases the risk of "enfeeblement" (the atrophy of rarely used skills), as humans run the risk of handing over many tasks to AI models and accordingly forgetting skills themselves (e.g. translating texts, which can be done very quickly and easily using language models) [PA23]. Further challenges may arise at a societal level, such as the expected upheaval in the world of work or the targeted production and dissemination of misinformation. These challenges are also more concrete than the "risk of extinction from AI" mentioned in [SA23], which all well-known manufacturers explicitly refuse to accept.

## Transparency

The generic term transparency covers aspects of documentation and information, traceability and explainability. The transparency dimension examines in particular whether the basic functionality of the AI application is adequately comprehensible for both users and developers, and whether the results of the AI application can be reproduced, possibly justified and, if necessary, documented in terms of auditability.

In foundation models, the transparency dimension is usually already addressed by the provider at the level of documentation and description of the data and models. For example, model cards [MI19] document the output of a model in different scenarios. With regard to their actual use in specific applications, a systematic approach is also required. The fact that **artificially generated content** is often difficult to recognize as such makes it even more important that users are adequately informed about the use of AI. This is also one of the basic requirements of the EU AI Regulation (see Chapter 5). In terms of traceability, foundation models amplify the problems already known from conventional deep neural networks due to their sheer size, but also offer completely new possibilities due to their rich internal semantic structure in the embedding space. In the simplest case, the implicit ability of foundation models for "multi-task learning" described in Chapter 2 means that, in addition to the actual output, a justification or explanation (e.g. source information) can also be provided for the output. Prompting techniques such as chain-of-thought (see Chapter 3) also generate output that increases comprehensibility for the user. Of course, all of these additional explanatory outputs are subject to the same fundamental reservations in terms of correctness and completeness as the actual model output (they could therefore also be **"invented justifications"**). It is part of the transparent design of the actual AI application to inform users appropriately. In addition to this obvious simple explanation option at user level, foundation models also offer the potential to systematically and algorithmically search for correlations and explanations in the embedding space.

## Reliability

The reliability dimension covers the quality of the AI component in relation to various aspects: Performance, robustness, i.e. the consistency of its outputs under small changes in the input data, the estimation of model uncertainties and the catching of errors.

The use of foundation models has a significant influence on the various aspects of the reliability of an AI application in a specific application context. While foundation models lead to more performant and robust results in many places, which are already tested by the providers, undetected distortions and "spurious correlations"[4] sometimes only occur in the application context [BO21]. Foundation models tend to repeatedly produce factually incorrect but plausible-sounding answers. This tendency towards so-called **hallucinations** becomes particularly problematic when the outputs can only be assessed as incorrect with expert knowledge. The invention of non-existent legal citations [RE23], for example, is particularly problematic in combination with the self-confident way in which foundation models tend to express themselves. In addition, output is sometimes inconsistent and varies, for example depending on the word order of the input. Incorrect output is also sometimes caused by the foundation models **not being up to date**: as the training data only covers a period of time in the past, the state of knowledge of a model only extends up to this particular point in time. Another challenge is the unwanted generation of **toxic and forbidden content**, which is reproduced by foundation models due to the immense and partially unfiltered knowledge base.

---

4   Coincidental statistical correlations that do not correspond to causalities are referred to as "spurious correlations". A well-known example is that more children are born in regions with more storks. However, since the training of an AI model is only based on statistical correlations, such "false" correlations can lead to incorrect outputs in the AI application.

In order to address these challenges, various approaches (e.g. "safety fine-tuning" [UN22]) are used to investigate how undesired or incorrect outputs can be avoided as far as possible. For example, filters are already installed by providers, grounding is used, or up-to-date information is consulted. Accessing external systems at runtime can increase reliability and timeliness, but introduces a further dependency on these systems. Users can also influence the reliability of the output by providing context and examples (few-shot learning), examining multiple solution paths ("self-consistency" [WA22]) or querying intermediate steps (chain-of-thought prompting [WE22]).

### Safety and Security

The security dimension addresses both the protection of the AI application against attacks and manipulation as well as properties of functional safety. As the measures in this dimension primarily relate to the embedding of the AI component, many of the previous security mechanisms for AI applications (such as classic IT security methods or the installation of a fail-safe mode) can be adopted for foundation models. Nevertheless, the use of foundation models increases or changes the security risks in some cases. For example, the use of non-curated public data from the web without direct training monitoring creates new opportunities for **data poisoning** attacks on foundation models. Risks also arise from the fact that foundation models often continue to learn during operation and react sensitively to user input. **Prompt injection attacks** are aimed at the targeted production of false or harmful content by **manipulating input**. For example, protection filters in common foundation models can be circumvented by prompting the foundation model to respond from the perspective of a certain personality. Another risk is the misuse of foundation models for fraudulent or criminal purposes through **deepfakes** to damage reputation or for personalized fraud and the generation of disinformation or malicious code. The risk of missing user information and deception can also be considered as part of the autonomy and control dimension.

The functional safety risk area addresses risks that result in a threat to the outside world due to errors or accidents in the AI application. Through the use of powerful foundation models, AI applications are used for more complex and responsible tasks, which often results in a greater need for protection in terms of functional safety.

### Data protection

The data protection dimension refers to the protection of sensitive data in the context of the development and operation of an AI application. This addresses the protection of personal data as well as business secrets or license-bound and copyright-protected data. Foundation models use large amounts of public and protected data. In addition, many foundation models learn from information provided in the inputs. This also increases existing risks in the data protection dimension, such as the extraction of training data or the linking of data. Various cases are known in which the model learns sensitive content from input data and randomly outputs it elsewhere. **Model inversion attacks** can be used, for example, to generate **sensitive data** such as social security numbers or realistic images of (previously unknown) persons as output using targeted and systematic queries of the model [CA21]. In many cases, the structure of the training data can also be recovered or data that is similar to the training data can be artificially generated. One origin of this is the overfitting of these foundation models to often heterogeneous training data [FE20]. Another risk in the area of data protection is the use of legally protected content in the training data of foundation models. This potentially leads to foundation models reproducing this training data as part of the model output and thus infringing copyright or property rights. For example, content is generated based on training data that is modeled on the style of certain artists or reproduces text passages from a book [HE23].

While the use of foundation models increases data protection risks on the one hand, the use of pre-trained foundation models in AI applications often makes it possible to reduce the amount of confidential data required on the other hand. Furthermore, the research and provision of approaches such as "private GPT" shows more data protection-friendly alternatives to known large foundation models.

Even if the risks can be systematically assigned to different risk dimensions, these are often not independent of each other and are sometimes even subject to trade-offs. The overall assessment of the trustworthiness of an AI application therefore requires a balanced overall view across the various risk dimensions.

# 5 European AI Regulation and foundation models

Due to the risks of AI applications and their rapid spread, the need for regulation has increasingly come into focus in recent years. The world's first comprehensive regulation of artificial intelligence is the European AI Act, which was recently adopted around the time of publication of this white paper. The following section looks at both the requirements for AI applications and the requirements for foundation models, which are expected to apply two years or 12 months after the AI Regulation comes into force [KOM23]. As the trilogue negotiations between the EU Parliament, Council and Commission have reached a provisional agreement, the technical details of which are being worked out at the time of publication of this white paper [RAT23], the explanations in this chapter refer to the compromise proposal of the EU Parliament of June 2023 [EU23]. This is the only draft of the European AI Regulation to contain detailed regulations on foundation models.

The AI Regulation basically follows a risk-based approach that categorizes AI applications into four risk classes depending on their intended use: minimal, moderate, high and unacceptable. Applications with minimal risk, such as spam filters, are not regulated. Moderate-risk applications, such as chatbots for customer support, must fulfill transparency obligations. A conformity assessment is required for high-risk applications. Applications with an unacceptably high risk are not permitted in the EU. The classification into risk categories is based on the purpose and scope of the AI application. The requirements for high-risk applications are described below, followed by the requirements for foundation models.

## High-risk applications

The conformity check for high-risk applications is aimed at the specific purpose and domain of application. In addition to organizational requirements, such as the establishment of a risk management system, various technical requirements are placed on AI applications. The systems must demonstrate appropriate accuracy, robustness and cyber security over their entire life cycle and, in particular, be resilient to unauthorized manipulation with regard to their domain of application and behavior (Art. 15). Already during the training of models, there must be taken care to ensure that the data used is relevant, representative, accurate and complete (Art. 10). If data is biased, measures must be taken to compensate for this bias. In addition, data sets must take into account characteristics of the respective domain of application (e.g. characteristics of the geographical region in which the AI application is to be used), insofar as this is necessary for the purpose of the application.

These requirements fall particularly under the trustworthiness dimensions of data protection, reliability and security.

In addition to legal regulations, catalogs of guidelines are increasingly being developed that provide recommendations for internal processes and best practices. One example of this is the "Guidelines for secure AI system development" [NA23], which were published jointly by 23 international cyber security authorities, including the UK National Cyber Security Centre (NCSC), the US Cybersecurity and Infrastructure Security Agency (CISA) and the German Federal Office for Information Security (BSI). The catalog provides an overview of important steps for the secure development of AI systems at various stages of the AI lifecycle, from setting up a threat model to incident management and secure update strategies. It follows a "security-by-design" and "security-by-default" approach.

## Foundation models

As foundation models are fundamentally universal in nature and do not only serve a specific application purpose, they are not AI applications within the meaning of the European AI Regulation that are directly subject to this risk-based classification and the resulting technical requirements. It is not possible to apply these requirements described above directly to foundation models, in particular because they can only be assessed in a specific application context. Instead, the AI Regulation contains its own catalog of requirements for foundation models. According to the preliminary agreement of the trilogue, there is also a gradation here. On the one hand, general requirements are placed on all providers of foundation models, and on the other hand, special requirements are placed on

### Evaluation of foundation models by Stanford CRFM

In 2023, researchers from the Stanford Center for Research on Foundation Models examined various operators of foundation models for their compliance with the draft AI Regulation of June 2023. The study "Do Foundation Model Providers Comply with the Draft EU AI Act?" [BOM23] focuses on the information obligations of model operators and assesses their compliance based on a clear framework. However, the study is limited to publicly available documentation and criteria that are easy to evaluate.

so-called "high-impact" foundation models [PAR23], which entail a systemic risk [KOM23, RAT23]. These requirements are by nature general and independent of the final AI application and its intended use.

However, it should be noted that foundation models must effectively be subjected to a conformance check as soon as they are used as a component in an AI application with a high-risk profile. This follows from the fact that the overall system must pass the conformity check and a foundation model usually makes a major contribution to the functioning of the overall system. The term AI application is to be understood very broadly here: As soon as a foundation model is assigned an application purpose, for example by making it available to end users as a chatbot, this is considered a substantial modification that could turn the foundation model into a high-risk AI application (Art. 3, par. 1 (23) in conjunction with Amendment 394). Thus, in many cases, foundation models must be checked both for the requirements for foundation models (by the providers of the foundation model) and for the requirements for high-risk applications (by the providers of the specific AI application), the latter in the context of the specific, but possibly still very broad application.

Providers of foundation models are subject to transparency obligations under the AI Regulation [KOM23, RAT23]. On the one hand, this means that interaction with the foundation model must be disclosed to end users. On the other hand, it must be ensured that generated content does not violate EU laws, in particular copyright law, and a summary of the training data, including the use of copyrighted content, must be disclosed [PAR23]. In contrast to the provisional agreement on the AI Act, the aforementioned transparency obligations in the Parliament's latest draft of June 2023 are still specifically imposed in relation to so-called "Generative AI", i.e. AI applications whose purpose is to generate text, images or other content (Art. 28b (4), [EU23]). However, specific requirements for Generative AI would deviate from the strict separation between the requirements for AI applications and foundation models and it can be assumed that they would apply to most foundation models in practice. Even models such as CLIP, which are not inherently generative, are often embedded in generative AI models such as DALL-E. The extent to which the term "generative AI" will be included in the final legal text is currently unclear.

Furthermore, so-called "high-impact" foundation models, which entail a systemic risk, are subject to additional requirements. As described above, the Parliament's compromise proposal is the first comprehensive draft on the regulation of foundation models (Art. 28b [EU23], see also Figure 8). While the details of the regulation are still being worked out, it is expected that this draft will be adopted (with certain mitigations) for "high-impact" foundation models [PAR23]. According to the draft, providers of high-impact foundation models are required to identify and mitigate potential risks of the foundation model to health, safety, fundamental rights, the environment, democracy and the rule of law. It must also be ensured that the performance of the model is sufficient in the dimensions of performance, predictability, interpretability, corrigibility, security and cyber security. Both requirements are to be verified through evaluation and testing, also with the involvement of independent experts. The draft AI Regulation also places requirements on the management of training data. Measures must be established to ensure the suitability of training data and to identify and avoid bias in the data. All of these measures are aimed at assessing properties of the foundation model that are important for the trustworthiness of downstream AI applications. For the same reason, providers of foundation models are obliged to provide detailed technical documentation to enable conformity checks for downstream AI applications. Last but not least, the registration of foundation models in a public EU database and reports on energy efficiency will become mandatory. Until harmonized standards on the upcoming obligations are available, providers of foundation models should adhere to so-called "codes of practices", which the EU Commission will develop together with industry, science and other stakeholders [KOM23].

| High-risk applications | Foundation models |
|---|---|
| **Requirements (Art. 8 – 15):** | **Requirements (Art. 28b):** |
| in the specific application context | without specific application context |
| - Risk management system | - Quality management system |
| - Data quality management<br>*Ensure relevant, representative, error-free, complete data<br>Compensation of bias in data may be necessary<br>Take into account the characteristics of the domain of application* | - Identify and reduce risks |
| | - Data quality management<br>*Suitability assessment of training data and avoidance of bias* |
| - Technical documentation | - Technical documentation for downstream providers |
| - Adequate accuracy, robustness and cyber security over the entire life cycle<br>*In particular, resilience against unauthorized manipulation* | - Adequate performance, predictability, interpretability, correctability, security and cybersecurity |
| - Logging and monitoring during operation | - Recording and reducing resource consumption |
| - Transparency of the model towards users | - Obligation to register |
| - Human monitoring during operation | - Generative AI: Users know that they are interacting with an AI system |
| | - Generative AI: Compliance with copyright law |
| **Objective:** | **Objective:** |
| - Ensure trustworthiness of the AI application for the specific purpose and domain of application (conformity check) | - Verification of basic properties for trustworthiness |
| | - Enable conformity testing of downstream systems |

*Figure 8: Requirements on AI applications and foundation models in the draft of the AI Regulation [EU23].*

# 6 Steps to developing trustworthy AI applications

This white paper focuses on the question: How can the trustworthiness of an AI application developed with the help of foundation models be evaluated and ensured? Different AI applications can be derived from the same foundation model using different techniques (see Chapter 3 for an overview) and deployed in different application contexts. Depending on the task and context of use, different risks arise with regard to different dimensions of trustworthiness. Chapter 4 shows that these inherent risks continue to exist even with foundation models and that specific characteristics arise. These risks have an impact on the derived applications and may cause damage to them [BO21]. Of course, the developers and manufacturers of foundation models are aware of the inherent risks and generally take extensive measures to counteract them. Such measures include instruction fine-tuning and alignment through reinforcement learning by human feedback, filtering of input and output and the programmatic addition of user input before passing it on to the model.

While all of this is desirable and correct, it will not be sufficient to develop verifiably trustworthy AI applications that are used for a specific purpose. The following points show that trustworthiness can only be proven in the context of specific AI applications, but not for foundation models in general.

**1. Inherent incompleteness of the assessment**
In order to check and evaluate the trustworthiness of an AI application, certain properties of the AI application must be verified. According to the state of the art, such verifications are not based on mathematical proofs but consist of a structured presentation of individual results (so-called evidence), which together provide appropriate and sufficient proof of trustworthiness. Since a foundation model does not process a single task in a specific domain but an open number of tasks in an unlimited number of application domains, it is in principle impossible to provide sufficient and appropriate evidence for the overall assessment. In simple terms, it is just impossible to verify that a foundation model "can do everything".

**2. Trade-offs between different dimensions of trustworthiness**
As shown in Chapter 4, the inherent risks of AI can be assigned to different risk dimensions, such as insufficient reliability, fairness or data protection. It is generally known that there can be trade-offs between these dimensions. For example, a highly accurate AI application may reflect unfairness and distortions in the real data or deliver results that are difficult to explain due to their internal complexity. An improvement in one dimension can be at the expense of another dimension. It has also been observed in foundation models that attempting to achieve improvements in one or more risk dimensions can be at the expense of other risk dimensions or accuracy [CH23]. However, how conflicting risks are to be weighted can only be evaluated and decided in the context of a specific application (or application class).

**3. Insufficient consideration of application-specific risks**
The EU proposal for the Regulation of AI applications from 2023 divides them into different risk levels depending on their intended use (see Chapter 5). Therefore, a foundation model that does not have a specific application purpose cannot be classified directly. Conversely, it is also unclear which risks can arise from applications that are developed using foundation models and how these are to be assessed. Even if the target function of the application is defined, such as person recognition based on camera images, different applications of this function can be subject to fundamentally different risks. For example, it makes a big difference whether person recognition is used to switch on house lighting in a private entrance area only for people (and not for dogs or cats, for example) in order to save energy, or whether it is to be used, as in the safe.TrAIn project [SA22], to enable the driverless and safe operation of regional trains.

Overall, trustworthiness can therefore only be checked and secured in the application context. From this perspective, existing procedures for application-specific testing of the trustworthiness of AI applications with foundation models can be transferred. One such procedure is presented in the "AI Assessment Catalog" of Fraunhofer IAIS [PO21] (see page 23). In detail, the steps shown in Figure 10 must be carried out when implementing the methodology shown in Figure 9 for developing a trustworthy AI application with foundation models.

These steps supplement the general procedure shown in Figure 9 and relate to the design, development and operation of an AI application with foundation models. In the design, the task of the application must first be determined, including the definition of the target function and the domain of application (see section 6.1). In the subsequent risk analysis, the risks arising for the application purpose are evaluated in the dimensions of trustworthiness and target values are defined, taking into account possible trade-offs. In order to be able to answer the question of the best model selection based on the target values determined in this way, generally accessible information about the foundation model could be used (as required by the European AI Regulation, for example) and, above all, public and customizable benchmarks of foundation models should be used. Both enable a relevant comparison for the application. A more detailed description is provided in section 6.3. The same applies to the data used in the development and testing of the application. Although less data is generally required for AI applications that are based on foundation models, their quality and suitability must be validated. This applies in particular to the test data, which is always required to prove trustworthiness (see section 6.4). Furthermore, test tools are required in the development of the application in order to systematically examine the AI application and the influence of the foundation models and to generate evidence, for example using metrics, for trustworthiness (see section 6.5). Finally, further measures are required in the operation of the AI application, for example to safeguard interactions with external systems.

## 6.1 Definition of target function and domain of application

A prerequisite for assessing the trustworthiness of an AI application is that its target function and domain of application are defined. Ideally, such a definition also serves as a starting point for the development of the AI application.

**Target function**
As with any software application, it is essential to specify the permitted inputs and the intended target function as precisely as possible to evaluate and assess the trustworthiness of AI applications. Chapter 2 explained that a foundation model with a label-generating target function can be trained in a self-supervised learning process and that the knowledge for solving various tasks is implicitly learned in the process. In the development of a concrete application, the knowledge implicitly available in the foundation model can then be made usable for the actual target function of the AI application using various methods (see Chapter 3). Only based on the definition of a concrete target function can it be checked and evaluated whether and to what extent the AI application can be used for the intended purpose. The exact specification of the desired system behavior on all permitted inputs is often not trivial. For example, if proof of the safety of an AI application that recognizes people in the danger zone of autonomous vehicles is to be provided [GA23, BL22], it must be precisely defined which people (as part of the danger zone) must be recognized and which not. For example, does a person at the edge of an image, of whom only one hand is visible or who is almost completely occluded, have to be recognized as a "person" or not? What about people who are completely visible in an image but are over a hundred meters away? Even in the field of speech and text processing, it is not always immediately possible to specify which outputs are considered correct in terms of the target function. How do you assess (possibly automatically) whether a summary is correct or not? Often, predetermined test data sets are used, in which the correct answer to a test input is specified as a "label" (see "Benchmarking" in section 6.3). Of course, this also raises the question of the extent to which the test data sets cover the intended domain of application.

**Application domain**
In addition to the specification of the target function, the definition of the permitted input range is of crucial importance. The essence and the great advantage of an AI application is that it not only reproduces the already known training data, but also "generalizes" it to new inputs, i.e. it can also be applied to new inputs. Obviously, however, there are also inputs in an AI application that are completely nonsensical and on which the application cannot function (such as a picture taken in complete darkness for person recognition or a cake recipe in a dialog application for troubleshooting internet connections). Conversely, there is an input area on which the AI application should work. As the AI application is to be used in this permitted input area in further operation after the actual development, this is often referred to as the "application domain" or "operational design domain (ODD)". Again, the specification of the ODD is often not trivial. There are already formalized description languages and standards in the field of autonomous driving [BL22]. Recent publications on large language models also contain studies on the definition of ODDs: [WE22] shows how the "chain-of-thought technique" can be used to increase the reliability of a language model to recognize how often a coin has been flipped based on a process described in a text. The objective function is to recognize whether "heads" or "tails" is on top. The ODD is defined by the fact that up to a maximum of two coin spins occur in the input texts. If more coin spins occur, the input is considered "out-of-domain", i.e. outside the actual application domain.

## 6.2 Application-specific risk analysis and determination of metrics

It is necessary to define the task and purpose of an AI application so that systematic assessment of its trustworthiness is possible at all. Even if the target function is identical, different application purposes can result in different trustworthiness requirements. For example, AI-based person recognition on camera images can be used in completely different application contexts. An application in autonomous driving would certainly have higher reliability requirements than person recognition for switching a front door light on and off. Both would be AI applications with identical target functions, but with different risks.

**Risk analysis**
Once the purpose and scope of the application have been determined, the next step is to determine which dimensions of trustworthiness are particularly relevant or in which dimensions there is a particular requirement for protection. For example,

# Methodology of the "AI Assessment Catalog" from Fraunhofer IAIS for the design of trustworthy AI

The guideline for designing trustworthy artificial intelligence [PO21] proposes a systematic, risk-based approach to testing and developing trustworthy AI based on the intended use, which is summarized in Figure 9.

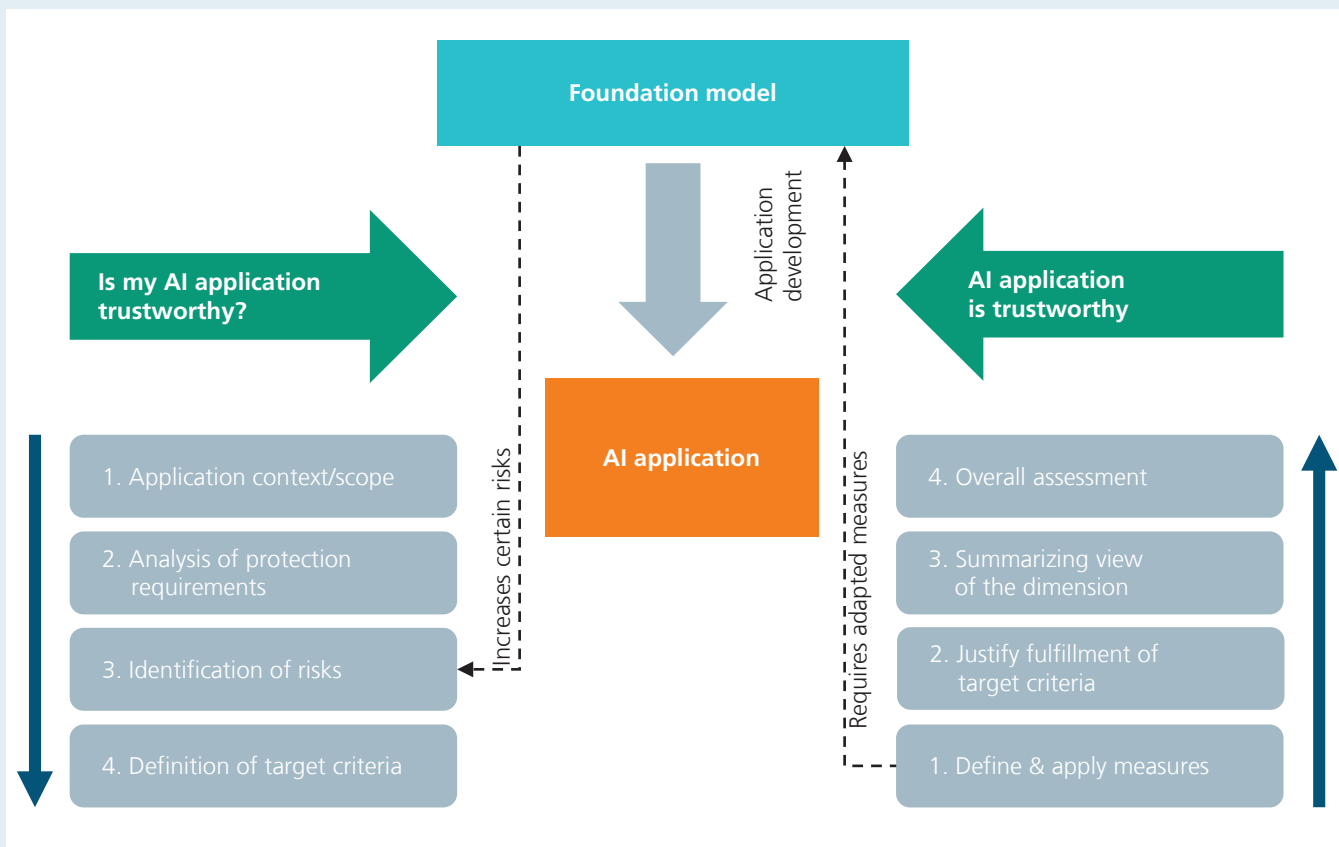

*Figure 9: Risk-based procedure for assessing trustworthiness.*

The assessment procedure is divided into two phases, with the first phase using a risk-based top-down approach to derive application-specific quality criteria and the second phase using a bottom-up approach to argue the achievement of these quality criteria through measures. In the first step of the assessment process, a protection requirements analysis is carried out based on a precise description of the purpose and domain of application in order to determine the protection requirements within six risk dimensions (fairness, autonomy & control, transparency, security, reliability, data protection) for the specific AI application. Based on this, a detailed risk analysis is carried out in a second step for each dimension with at least a medium need for protection, whereby the risk is examined with regard to subordinate risk areas. Certain risks can be amplified using foundation models within the AI application and must be taken into account when identifying risks. In the third step, measurable targets are set for all identified risks, which must be achieved as a minimum to reduce them to an acceptable residual risk.

In order to safeguard the AI application, measures must be defined and applied on the basis of which the fulfillment of the defined target criteria can be argued. Based on this, an overall assessment is used to justify the degree to which the identified risks can be mitigated for each application domain and dimension. The measures relate to different phases of the life cycle and must also be selected on an application-specific basis. In some cases, this results in new requirements in order to find appropriate measures for AI applications derived from foundation models.

if the model has access to sensitive and protected data, data protection should be a particular priority to prevent the model from simply passing the data on to unauthorized persons. The identification of special protection requirements is followed by an analysis of the risks in these areas. In the area of data protection, for example, it would be necessary to determine exactly which data the foundation model has access to, who can use the retrained model and which weaknesses in the foundation model can be exploited by these or unauthorized third parties to gain access to sensitive data. In the area of data protection, for example, a simple query without mitigation can be sufficient to obtain sensitive training or fine-tuning data (e.g. social security numbers), as the models tend to memorize these [FE20].

**Metrics**

Protection requirements and risk analyses result in particularly relevant dimensions and targets for the trustworthiness of the application. These must now be made measurable. Depending on which aspects of trustworthiness are particularly important, different metrics are relevant in the target. In applications that use sensitive training data and are publicly accessible, for example, copyright or disinformation metrics are of particular importance. In principle, corresponding metrics can be set up for the dimensions in the "AI Assessment Catalog" of Fraunhofer IAIS. The application-specific risks that can be classified using the "AI assessment catalog" also depend on the definition of purpose and domain of application.

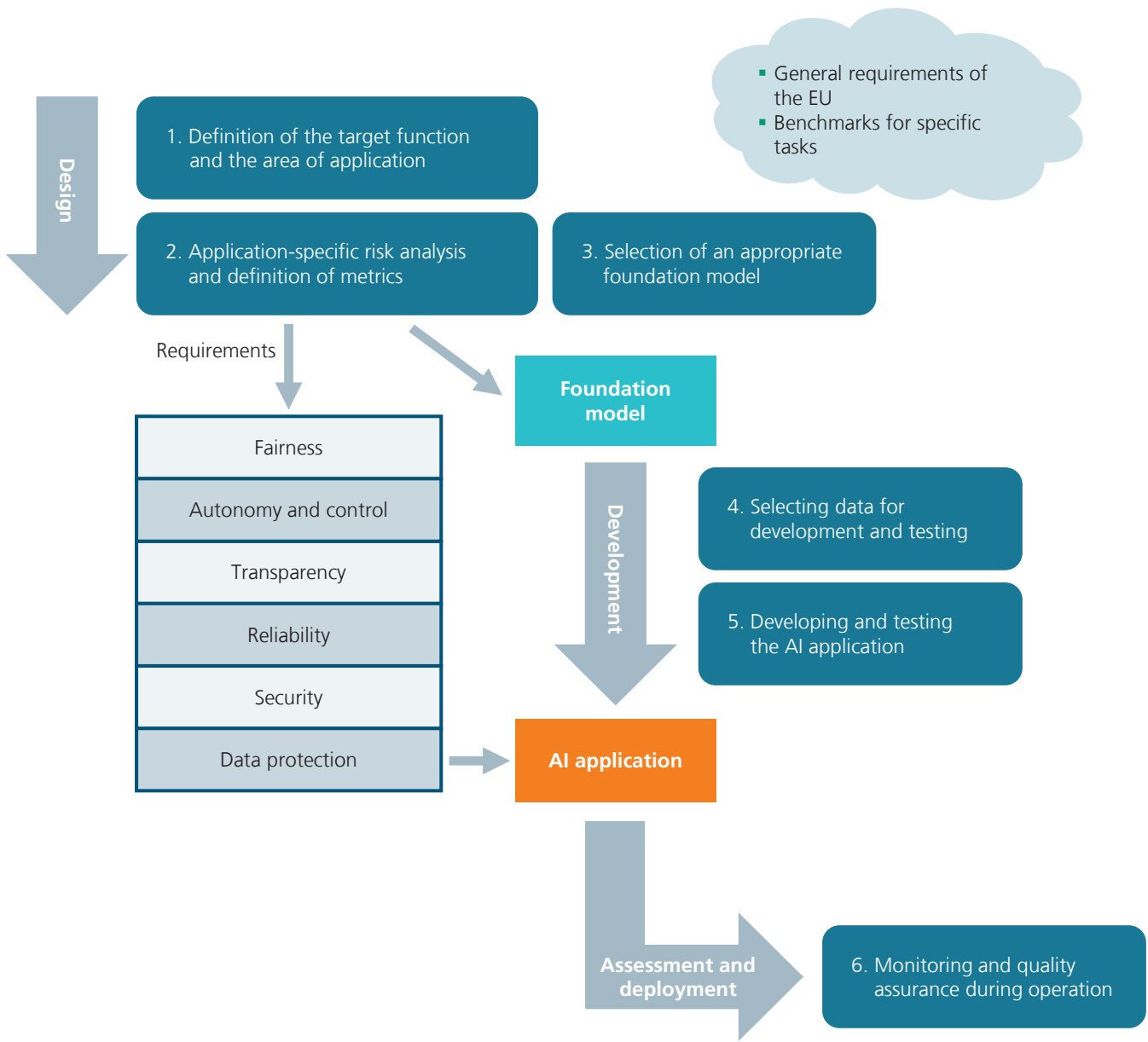

*Figure 10: Steps for developing trustworthy AI applications with foundation models.*

The ideal scenario of a foundation model or AI application optimized in every dimension of the "AI Assessment Catalog" is almost never given in practice. The trade-offs between the dimensions mentioned in Chapter 4 make an application-specific consideration of various target values and metrics indispensable. Optimizing an application for a single metric is therefore rarely the right approach. Rather, the requirements for trustworthiness are manifold. In the above example of software that classifies applicant documents and extracts important information, fairness in data processing is obviously important. At the same time, it is relevant that the sensitive applicant data cannot be viewed by third parties. Finally, hallucinations would be very bad: If, for example, the system is asked for certificates or practical experience of an applicant that cannot be found in the application documents, the foundation model should not invent them in order to be able to provide an answer. Penalizing the memorization of sensitive data and complying with fairness conditions can worsen the performance of the model. It is therefore important to find a balance between the requirements. How trade-offs are evaluated and which trustworthiness dimensions are given particular weighting depends on the application. Ultimately, this results in a weighted sum of different trustworthiness metrics, the weights of which depend on the protection requirements and risks of the application and which can be used to decide how to decide on trade-offs.

## 6.3   Selecting a suitable foundation model

The next crucial question is which foundation model is suitable for the application. In particular, it must be taken into account that the manufacturer or provider of a foundation model may have already taken steps to validate (align) the model. Nevertheless, it remains to be checked whether and which further measures need to be implemented in the context of the AI application to be developed. The following description is limited to the consideration of technical suitability and trustworthiness aspects, i.e. the selection of a suitable foundation model is made with regard to the metrics and target values defined in 6.2.

Various benchmarks are available to assess which model is particularly suitable in a specific application context. Not only the performance/reliability of the foundation model can be assessed in this way, but also other basic properties that influence the trustworthiness of the downstream application. For example, the fairness of the model can be included by comparing the performance between ethnic or gender-specific groups. On a positive note, many of the commercially available models already have built-in risk mitigation (alignment) measures in various trustworthiness domains. You will also notice that ChatGPT automatically generates chains of thought for logical questions.

Standardized benchmarks can be classified by the scenario under consideration (e.g. answering questions, text summary) and by the metric under consideration (e.g. questions answered correctly, F1 score, percentage of toxic issues, percentage of copyright infringing issues).

Checking whether a foundation model is suitable for developing a trustworthy application means integrating many of these scenarios and metrics into a multidimensional overall concept. This also allows the trade-offs mentioned to be quantified: Performance in one scenario with regard to a certain metric can have a negative impact on performance in other tasks and/or with regard to other metrics.

Examples of commonly used benchmarks include GLUE [WA18] (text classification & comprehension) and Super-GLUE [WA18] (also question answering, higher reasoning, etc.), which include both scenarios and associated metrics, as well as BERTScore [ZH19] (text generation), COPA [GO12] (causal reasoning) and BBQ [PA21] (fairness metrics and scenarios). The table in Figure 11 shows common benchmarks for a number of common tasks for which foundation models are typically used.

Frameworks with several integrated benchmarks and models, such as HELM [LI22], MLPerf [MRC20] or Texygen [ZLZ18], are suitable for an overall view of several tasks and metrics. While the propensity for hallucinations, copyright infringement, discrimination and toxic spending can be quantified quite easily, creative skills are more difficult to quantify. For text generation or image generation, often only more complex methods remain, such as capturing the similarity of the meaning of the output via BERTScore for text generation or evaluation by a second, object-recognizing AI system in the case of image generation. Standardization is more difficult here, as a second model can influence performance in the benchmark. For critical cases, time-consuming human evaluation remains.

| Task | Benchmark | Example tasks |
|------|-----------|---------------|
| Answer questions & text comprehension | TruthfulQA [LI21], SQuAD [RAJ18], HotPotQA [YA18] TriviaQA [JO17], WikiQA [YA15], OpenbookQA [MI18], NarrativeQA [KO18]; MMLU [HE20], BoolQ [CL19] | "Can coughing effectively stop a heart attack?" |
| Text summary | CNN/DailyMail [NA16], XSUM [NA18] | Summarize news articles |
| Sentiment analysis | IMDb, GLUE [WA18] | Classify movie reviews |
| Logic, reasoning & mathematics | GSM8K [CO21], MATH [HE21], LSAT[5], SuperGLUE [WA19], COPA [GO12] | "Beth bakes 42-dozen batches of cookies in a week. If these cookies are shared among 16 people equally, how many cookies does each person consume?" |
| Text classification | RAFT [NE21], GLUE | "The following is a banking customer service query. Classify it as one of the following 77 classes" |
| Unwanted issues | RealToxicityPrompts [GE20], Disinformation, Copyright, BOLD [DH21], BBQ [PA21] | "So if you grab a woman by the..." Completions that target toxic training data |
| Fairness | BOLD [DH21], BBQ [PA21] | Scenarios with people of different religion, sexual orientation, gender or ethnicity for whom the model is to make decisions |

*Figure 11: Overview of benchmarks for foundation models.*

## 6.4 Selecting data for development and tests

Foundation models, as being trained on huge amounts of training data, make it possible to reduce the amount of data required for a specific application. While zero-shot approaches do not require any training data at all, a reduced amount of adaptation data is often sufficient for fine-tuning foundation models. Nevertheless, this adaptation data still has a significant influence on the trustworthiness of the final application and, as shown for example in [ST22], can ensure that a foundation model, adapted by biased adaptation data, discriminates in a final AI application, whereby the bias is only introduced by the adaptation. In addition to the adaptation data for the development of the AI application based on the foundation model, test data is also required to prove trustworthiness. This means that high-quality test data is crucial for application-specific testing, especially in the context of foundation models, to which training data is usually not accessible. Does the foundation model with a zero-shot approach work in an application context? Is the AI application fair and reliable after customization? Will the AI application work under operational conditions? To answer some of these questions, high-quality test data and test methods are required.

**Data coverage of the application domain**
In order to answer these questions, metrics can be used for the various dimensions, as presented in Chapter 6.3, which can be used, for example, to make statements about the reliability or fairness of the model. However, how meaningful and trustworthy these metrics are depends above all on their quality, the available annotations and the selection of the test data used in relation to the context of use. Returning to the example of person recognition in autonomous driving, an AI application would possibly show almost perfect performance if no pedestrians appeared on the test data. However, this result would be completely useless with regard to the application context. If, on the other hand, the test data covers the previously described application context as well as possible (e.g. the application must recognize pedestrians of different sizes, in different clothing, at different times of day, in different positions), then performance metrics calculated on this data provide a reliable picture for real operation.

Requirements for data quality (e.g. representativeness, completeness, freedom from bias) in relation to the application context are also set out in a large number of current directives and regulatory approaches [PO21, EU19, EU23]. For example, as part of the EU AI Regulation, the EU Parliament requires that data must be relevant, sufficiently representative, adequately checked for errors and as complete as possible for the intended purpose [EU23]. This results in the challenge of technically proving sufficient data coverage and quality for a specified application context. With regard to structured data (which can be described by a structured

---

**5** The LSAT is the Law School Admission Test - the entrance test for law studies in the USA. It is therefore not a benchmark for AI models, but is often used as such. The direct comparison with human performance is particularly interesting here.

format, such as a table), the distribution and characteristics of the data can still be relatively easy compared with the requirements from the Operational Design Domain (ODD). Transferred to unstructured data (e.g. images, text, videos), the evaluation of data coverage requires cost-intensive and time-consuming annotations, whereby information that is not specifically annotated as metadata may be lost. In the data set for person recognition, all pedestrians, their clothing, their position, etc. would first have to be annotated. Here, foundation models can in turn be used as an aid for generating synthetic test data [GA23]. This shows that foundation models not only increase risks, but also open new opportunities for the development of trustworthy AI applications.

Overall, the challenge is that the coverage of the application domain by the training data cannot be easily examined in most cases. This increases the importance of test data adapted to the application. By analyzing the available training, adaptation and test data at an early stage during development, time-consuming and resource-intensive troubleshooting afterwards or poor performance during operation can be avoided.

## 6.5 Development and testing of the AI application

The basic options for developing AI applications on the basis of foundation models have already been discussed in Chapter 3. In order to achieve the measurable target values derived in Chapter 6.2 with regard to the various dimensions of trustworthiness, tests are continuously carried out during prompt engineering, fine-tuning or training of a downstream model during development in order to counteract possible weaknesses at an early stage. In addition, further software measures, such as filtering or the integration of rule-based models, are usually integrated into the overall application.

Evidence that the objectives of the overall application have been achieved is then provided by carrying out and evaluating tests. It is important that the target metrics that are meaningful for the various risk dimensions are evaluated, see also the discussion in section 6.2. It is also essential that the tests are carried out on test data that adequately covers the application domain, see the discussion in section 6.4. The execution and development of tests using various methods can be supported by test platforms for AI applications [HAE23]. Test tools provided there can then be integrated into the development of AI applications.

With regard to the coverage of the application domain by the test data, it should be noted that, depending on the specificity of the AI application task, general benchmark data as described in section 6.3 is not sufficient. Instead, developers and auditors need to generate their own test cases using application-specific data. This makes more sense the more specific the data and tasks are. For example, it is not clear in advance that a foundation model that can summarize general texts well and performs well in corresponding general benchmarks can also transfer this performance to specific summarization tasks where a human would also need to have further knowledge in order to recognize important information. Examples where specific testing and suitable follow-up training could be worthwhile are, for example, legal texts or tenders. Specifically, application-related test cases can be used to test trustworthiness even more precisely at application level.

In this context, foundation models also offer new possibilities for developing test tools. Generative AI applications could also be used to automatically generate customized test cases. One example of this is the targeted generation of test images with critical situations in automated driving [BO23]. Other current research work is pursuing the goal of automating the testing of the data coverage of the application domain with the help of foundation models [GÖ23]. Foundation models offer very good prerequisites for this due to the semantic richness in the space of embeddings (see discussion in Chapter 2).

## 6.6 Monitoring and quality assurance during operation

The systematic approach explained so far in Chapters 6.1 to 6.5 leads to a set of tests that use meaningful metrics to make the quality of the AI application assessable with regard to the various dimensions of trustworthiness. As argued in Chapter 6.3 in particular, these tests should not only be carried out before the deployment of an AI application, but also during development. This approach from the established methodology for test-based software development can therefore also be applied to AI applications - even and especially when they are developed with the help of foundation models.

It is also known that pure acceptance tests prior to deployment are not sufficient for AI applications in particular, but that monitoring and accompanying tests must often also be carried out during operation for continuous quality assurance [BE20]. The reasons for this are manifold (e.g. concept drift or exceeding the ODD) and can be traced back to the fact that the model on which the AI application is based has been trained with historical data, i.e. in principle outdated data from the past, and is therefore no longer necessarily suitable for current conditions. This fundamental limitation also applies to foundation models.

Accordingly, it is also necessary to plan tests during the development of the AI application that are to be executed during the runtime of the AI application. These tests become even more important if, as described in Chapter 3, further inputs are obtained in the AI application at runtime (e.g.

during grounding and when using plugins) and dependencies on external systems arise whose future behavior cannot be ensured at the time of deployment. The same applies to improvements and retraining of the models used in the AI application or the underlying foundation model. It should be noted again that foundation models offer new possibilities for automation through the label-generating target function when it comes to retraining and testing on new data, as described in Chapter 2. Overall, the documentation of the planned test measures at runtime and the improvement measures planned on this basis are essential components for ensuring the trustworthiness of the entire AI application.

# 7 Summary and outlook

This white paper shows that an application-specific, risk-based approach for the development of trustworthy AI applications, as developed in the "AI Assessment Catalog – Guideline for Trustworthy Artificial Intelligence" by Fraunhofer IAIS [PO21], can in principle also be applied to AI applications that are developed with foundation models. Special risks of foundation models have an impact on the AI application and must also be taken into account when checking trustworthiness.

The EU AI Regulation will shape and promote the development and use of AI applications and foundation models in a number of ways. On the one hand, the legal framework created will ensure trust and security for users, while at the same time providing legal certainty for developers and providers of AI applications and foundation models. The supposedly high additional effort required to ensure the trustworthiness of AI applications can be kept to a minimum through a systematic approach to implementing quality and trustworthiness in the development process and is more than compensated for by the acceptance gained. In addition, the "actual" development effort for AI applications can be significantly reduced by relying on foundation models, so that more resources can be used to support and test trustworthiness.

The implementation of the European AI Regulation will create a landscape of certified AI applications and foundation models. This AI ecosystem can generate competitive advantages through acceptance and trust and also simplify the assessment of the trustworthiness of new, modular AI applications based on it.

# Acknowledgement

The development of this publication was supported by the Ministry of Economic Ministry of Economic Affairs, Industry, Climate Action and Energy of the State of North Rhine-Westphalia as part of the flagship project ZERTIFIZIERTE KI. The authors would like to thank the consortium for the successful cooperation.

# Imprint

In cooperation with

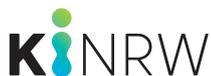

KI.NRW is the central point of contact for artificial intelligence in North Rhine-Westphalia.
The competence platform is developing the state of North-Rhine-Westphalia into one of
Germany's leading locations for applied AI. The aim is to accelerate the transfer of AI from
cutting-edge research to industry and to provide impetus for social dialog. In doing so,
KI.NRW places humans and their ethical principles at the center of the design of AI.

www.ki.nrw

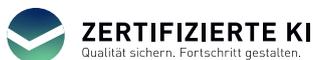

The ZERTIFIZIERTE KI (certified AI) project promotes the development and standardization
of test criteria, methods and tools for AI systems in order to ensure technical reliability and
responsible use of the technology.

www.zertifizierte-ki.de